\documentclass{article}

\PassOptionsToPackage{numbers,sort&compress}{natbib}
\usepackage[preprint]{neurips_2025}

\usepackage{booktabs}
\usepackage{multirow}
\usepackage{makecell}
\usepackage{threeparttable}
\usepackage{tabularx}
\usepackage{wrapfig}

\usepackage{amsmath}
\usepackage{amssymb}
\usepackage{dsfont}  

\usepackage{tikz}
\usetikzlibrary{shapes,arrows,arrows.meta,positioning,calc,intersections,backgrounds,fit}

\usepackage{xspace}
\newcommand{\ours}{Labimus\xspace}

\providecommand{\eg}{\textit{e.g.}\xspace}

\usepackage{subcaption}

\usepackage{pifont}
\usepackage{xcolor}
\newcommand{\cmark}{\textcolor{green!60!black}{\ding{51}}}  
\newcommand{\pmark}{\textcolor{gray}{\ding{51}}}            
\newcommand{\xmark}{\textcolor{red!70!black}{\ding{55}}}    

\usepackage[breaklinks,colorlinks,citecolor=blue,urlcolor=blue,linkcolor=blue]{hyperref}

\title{Labimus: A Simulation and Benchmark for Humanoid Dexterous Manipulation in \\
Chemical Laboratory}

\author{
Yuhan Wu\textsuperscript{1*}\quad
Zhao Jin\textsuperscript{2*}\quad
Tao Li\textsuperscript{2}\quad
Yuheng Zhang\textsuperscript{2}\quad
Zhichao Wang\textsuperscript{1}\quad
Shuo Wang\textsuperscript{1}\\
\bfseries Jun Jiang\textsuperscript{1}\quad
Xiaobo Li\textsuperscript{1$\dagger$}\quad
Yanyong Zhang\textsuperscript{1$\dagger$}\quad
Jian Tang\textsuperscript{2$\dagger$}\quad
Zhengping Che\textsuperscript{2$\dagger$}\quad
Yan Xia\textsuperscript{1$\dagger$}\\[0.5em]
{\mdseries\textsuperscript{1}University of Science and Technology of China\quad
\textsuperscript{2}Beijing Innovation Center of Humanoid Robotics}\\[0.3em]
{\mdseries\url{https://labimus.github.io/}}
}

\begin{document}
\maketitle
\renewcommand{\thefootnote}{\fnsymbol{footnote}}
\footnotetext[1]{Equal contribution.}
\footnotetext[2]{\raggedright Corresponding authors. yan.xia@ustc.edu.cn, xiaoboli@ustc.edu.cn, z.che@x-humanoid.com, jian.tang@x-humanoid.com, yanyongz@ustc.edu.cn\par}
\renewcommand{\thefootnote}{\arabic{footnote}}
\vspace{-3em}
\begin{figure}[h!]
\centering
\includegraphics[width=\textwidth]{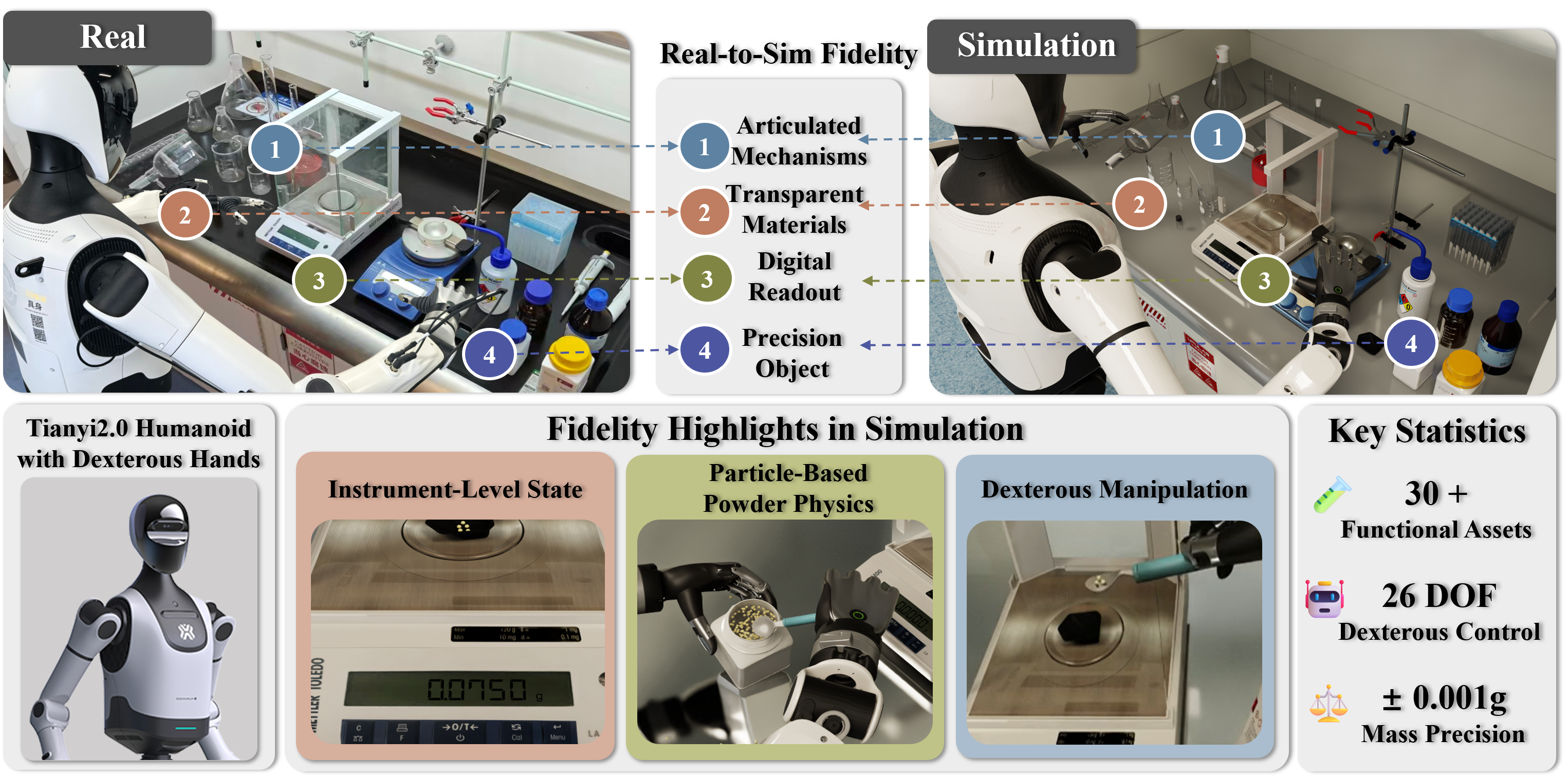}
\caption{\textbf{Overview of \ours.}
\emph{Top}: real-to-sim reconstruction of a chemistry workstation with 30+ functional assets covering the fundamental operations of organic chemistry experiments.
\emph{Bottom}: Tianyi\,2.0 humanoid performs precision-critical operations, showcasing instrument-level state readouts, particle-based powder physics, and contact-rich dexterous manipulation.}
\label{fig:overview}
\end{figure}

\begin{abstract}
    
    Laboratory automation has made remarkable progress through robotic platforms and AI-driven scientific reasoning. However, many laboratory operations (\eg, solid--solid transfer) remain inherently dynamic and require real-time adaptation to different materials and experimental conditions. Such precision-critical manipulations are difficult to standardize, motivating the use of humanoid robots with dexterous hands. Despite this opportunity, no existing benchmark evaluates humanoid manipulation in precision-critical laboratory environments.
    We present \ours, to our knowledge the first benchmark for humanoid dexterous manipulation in organic chemistry laboratories. \ours reconstructs over 30 functionally faithful assets from real organic chemistry workstations through real-to-sim modeling, collectively covering the core operations of routine organic chemistry experiments. The benchmark integrates articulated laboratory instruments, particle-based powder physics, and closed-loop instrument readouts, enabling a complete manipulation-to-measurement pipeline. It further defines six atomic operations and a seven-step solid-weighing workflow derived from real laboratory standard operating procedures. We introduce a precision-aware evaluation protocol designed to jointly measure task completion, experimental precision, and long-horizon execution.
    We benchmark three representative policies under procedural layouts and environmental perturbations. Results reveal a precision gap: policies that successfully complete laboratory tasks can still fail to satisfy the quantitative tolerances required by experimental protocols. Our benchmark exposes a fundamental disconnect between task completion and experimental validity, providing a new testbed for developing reliable humanoid robots for scientific laboratories.
    
    \end{abstract}

\section{Introduction}
\label{sec:intro}

Laboratory automation has made remarkable progress toward autonomous scientific experimentation. Robotic platforms enable high-throughput and reproducible experiments~\cite{mobile_robotic_chemist,labutopia}, while AI agents assist with experimental design, protocol planning, result interpretation, and iterative decision-making~\cite{coscientist,chemcrow}. Most existing systems follow a common paradigm. They customize laboratory layouts and redesign tools so that fixed robotic arms with task-specific grippers can perform predefined experiments.
However, many laboratory operations are inherently dynamic and cannot be standardized.
Consider solid--solid transfer in organic chemistry.
A chemist must adjust the scooping force according to the powder's granularity and cohesion.
The powder is then deposited into a weighing boat and placed at the center of an analytical balance.
The entire process requires milligram-level precision.
It also demands multi-finger coordination, fine force control, and sustained tool-mediated contact.
Such operations are difficult to automate with customized fixed-arm systems.
Moreover, this paradigm is confined to purpose-built facilities and cannot be easily deployed in ordinary laboratories.
Humanoid robots with dexterous hands offer a more general solution.
They can operate directly in existing laboratories using standard tools and instruments, without redesigning either the environment or the hardware.

Despite this opportunity, humanoid dexterous manipulation for precision-critical laboratory tasks remains unexplored.
Existing humanoid manipulation benchmarks~\cite{geniesim3,behavior1k} and dexterous hand benchmarks~\cite{maniskill3} focus on household tasks, such as picking up mugs or opening drawers.
These tasks are evaluated with binary success criteria.
An object is either grasped or not, and a drawer is either open or closed.
Laboratory manipulation is fundamentally different.
For example, transferring exactly 50\,mg of powder is far more than a pick-and-place task.
The robot must regulate contact force throughout scooping, transporting, and depositing.
Success depends on the final mass error, not simply on whether the transfer is completed.
Existing household benchmarks do not evaluate such precision-critical manipulation.
Meanwhile, laboratory simulation platforms~\cite{labutopia,chemistry3d,autobio,matterix} have advanced scene construction and chemical process modeling.
However, they all adopt fixed robotic arms with parallel-jaw grippers.
Humanoid dexterous manipulation is not considered.
As a result, no existing benchmark combines humanoid dexterous hands, precision-critical laboratory tasks, and quantitative evaluation.

Building such a benchmark is challenging.
First, laboratory simulation requires more than rigid-body dynamics.
Powders behave as granular media and require particle-level simulation.
Laboratory instruments also require articulated mechanisms and real-time state readouts to produce physically meaningful measurements.
Without these components, experimental outcomes cannot be verified.
Second, laboratory manipulation is contact-rich and tool-mediated.
Tasks such as scooping powder or sliding a glass door require coordinated finger motion and precise force control, far beyond simple pick-and-place.
Third, evaluation should measure experimental quality rather than binary success.
Continuous precision metrics are needed to determine whether the manipulation satisfies experimental tolerances (e.g., $\pm$0.001\,g).
The intersection of faithful laboratory simulation, humanoid dexterous manipulation, and precision-aware evaluation therefore remains entirely unexplored.

To address this gap, we present \ours (Figure~\ref{fig:overview}), to our knowledge the first benchmark for humanoid dexterous manipulation in precision-critical laboratory environments.
\ours is built from real organic chemistry workstations through real-to-sim reconstruction.
It contains over 30 functionally faithful assets that collectively cover the core operations of organic chemistry experiments.
The assets include flasks, beakers, spatulas, analytical balances, and other commonly used laboratory instruments.
Each asset supports articulated mechanisms, particle-based powder physics, or closed-loop instrument readouts (e.g., real-time digital mass display), depending on its functionality.
Together, they establish a complete loop from physical manipulation to quantitative measurement.

Based on these assets, \ours defines six atomic operations (door open, door close, grasp \& place, tare press, tool pickup, and scoop \& weigh) together with a complete seven-step solid-weighing workflow derived from standard operating procedures (SOPs) in real organic chemistry experiments.
Evaluation follows a three-tier hierarchy.
We measure binary task completion, quantitative precision against protocol tolerances (e.g., $\pm$0.001\,g), and long-horizon execution through step-level progress tracking and stage-level diagnostics.
Every task is evaluated under procedural layouts and three perturbation conditions: lighting, texture, and their combination.
Together, they form a $3 \times 4$ evaluation matrix.
We benchmark three representative policies: ACT~\cite{act_ref}, Diffusion Policy~\cite{dp_ref}, and $\pi_0$~\cite{pi0_ref}.
Across all methods, we observe a clear finding.
Policies that complete laboratory tasks successfully may still fail to satisfy the quantitative tolerances required by experimental protocols.
Our results reveal a fundamental disconnect between task completion and experimental validity, highlighting an important challenge for future embodied AI in scientific laboratories.

To summarize, our core contributions include:

\begin{itemize}

    \setlength\itemsep{0.15em}

    \item \textbf{The first benchmark for humanoid dexterous manipulation in precision-critical chemical laboratories.} We introduce the first benchmark dedicated to humanoid dexterous manipulation in organic chemistry laboratories. It covers six atomic operations and a complete seven-step solid-weighing workflow derived from real laboratory SOPs.

    \item \textbf{A high-fidelity laboratory simulation for humanoid dexterous manipulation.} We reconstruct over 30 functionally faithful assets that collectively cover the core operations of organic chemistry experiments. The simulation integrates articulated laboratory instruments, particle-based powder physics, and closed-loop instrument readouts, providing realistic interaction and measurement for embodied agents.
    
    \item \textbf{A precision-aware evaluation protocol for humanoid laboratory manipulation.} We introduce a multi-level evaluation framework that jointly measures task completion, experimental precision, and long-horizon execution. Experiments reveal a gap between successful task completion and experimentally valid manipulation, highlighting an important challenge for future humanoid laboratory robots.
    
\end{itemize}

\section{Related Work}
\label{sec:related}

\paragraph{Manipulation benchmarks.}
Large-scale simulation benchmarks have driven rapid progress in robotic manipulation.
RLBench~\cite{rlbench}, CALVIN~\cite{calvin}, RoboCasa~\cite{robocasa}, and LIBERO~\cite{libero} collectively provide hundreds of tasks spanning pushing, grasping, door opening, and object rearrangement.
However, all employ a single-arm parallel-jaw gripper and evaluate binary task completion.
Habitat~3.0~\cite{habitat3} extends embodied AI evaluation to social and collaborative scenarios but does not address precision manipulation.
RoboGen~\cite{robogen} automates task generation via generative simulation.
RoboTwin~2~\cite{robotwin2} introduces 50 dual-arm tasks across five embodiments, but the tasks are predominantly tabletop bimanual manipulation without sustained contact-rich laboratory operations.
BEHAVIOR-1K~\cite{behavior1k} scales to 1{,}000 household activities with humanoid support, but its tasks center on domestic scenarios and use binary completion metrics.
Factory~\cite{factory} and FurnitureBench~\cite{furniturebench} are the closest to precision manipulation.
Factory defines contact-rich industrial assembly tasks in Isaac Gym~\cite{isaacgym}, and FurnitureBench targets long-horizon furniture assembly.
However, both are limited to single-arm setups without laboratory instrumentation.
None of these benchmarks operates in a laboratory setting such as chemistry or biology, nor measures per-step execution precision.

\paragraph{Humanoid and dexterous manipulation.}
Recent benchmarks have begun to explore humanoid embodiments and dexterous end effectors, but their tasks remain far from the precision-critical manipulation required in chemistry and other laboratory sciences.
Dexterous manipulation research has produced strong results in in-hand manipulation~\cite{openai_dactyl,openai_rubik}, learning from demonstrations~\cite{dapg,dextransfer}, and dexterous grasping and articulated-object manipulation~\cite{dexgraspnet,unidexgrasp,dexart,gapartnet}.
However, none of these advances has been applied to laboratory task benchmarks where quantitative precision matters.
Genie Sim~3.0~\cite{geniesim3} builds a comprehensive humanoid simulation platform with LLM-driven scene generation, over 5{,}000 object assets, and 200 representative tasks.
However, its benchmark suites evaluate instruction following, spatial reasoning, and manipulation skills via binary task success, without measuring quantitative execution precision against laboratory tolerances.
ManiSkill3~\cite{maniskill3} offers GPU-parallelized simulation across 12 task domains and includes both humanoid and dexterous-hand categories.
Yet its dexterous tasks center on in-hand rotation and finger manipulation, and its humanoid tasks focus on pick-and-place rather than contact-rich, precision-critical laboratory operations.
DexJoCo~\cite{dexjoco} designs tasks that highlight the unique capabilities of dexterous hands over parallel grippers.
However, it targets everyday activities rather than laboratory procedures, uses a fixed-arm setup, and evaluates only binary task success.
In the broader humanoid learning community, recent works on whole-body control~\cite{humanplus,h2o,omnih2o} and humanoid benchmarks~\cite{humanoidbench} have advanced locomotion and manipulation capabilities, including dexterous hand teleoperation.
However, their tasks target general-purpose scenarios rather than precision-critical laboratory manipulation with quantitative tolerances.
In summary, existing humanoid and dexterous benchmarks evaluate either coarse household manipulation with humanoid bodies or isolated fine-motor skills with dexterous hands on fixed arms.
Even dexterous benchmarks that explicitly design tasks around hand-specific capabilities~\cite{dexjoco} remain in general-purpose domains without a humanoid embodiment or continuous precision evaluation.
The intersection of humanoid dexterous manipulation and precision-critical laboratory tasks remains entirely unaddressed.

\paragraph{Scientific experiment simulation.}
Several recent platforms bring laboratory settings into simulation, advancing scene construction, reaction modeling, and workflow-level automation.
LabUtopia~\cite{labutopia} introduces chemistry-aware physics simulation with GPU-accelerated fluid dynamics and a reaction-modeling engine covering 200 substances, organizing tasks into a five-level hierarchy.
Its Level-4 experiments reveal catastrophic success-rate decay across multi-step sequences (from 99\% at the first sub-step to under 2\% at the seventh).
However, LabUtopia uses a single Franka arm with a parallel-jaw gripper (neither humanoid nor dexterous) and evaluates all tasks with binary success rate.
Chemistry3D~\cite{chemistry3d} provides a toolkit with real-time reaction visualization, but manipulation tasks are demonstrated qualitatively without standardized evaluation.
AutoBio~\cite{autobio} defines 16 biological laboratory tasks and evaluates representative VLA baselines, finding that precision failures compound in multi-step tasks.
It introduces custom physics plugins for threaded assemblies and quasi-static liquids.
Notably, although AutoBio includes a 19-DOF hand, most joints are fixed to emulate gripper-like operation; full dexterous support is deferred to future work.
MATTERIX~\cite{matterix} constructs a GPU-accelerated digital-twin framework for chemistry workflows but positions itself as a development platform rather than a benchmark.
In summary, existing laboratory simulation platforms have advanced reaction modeling and scene construction for chemistry and biology from complementary directions, but all effectively operate with fixed-arm, gripper-like configurations.
None evaluates humanoid dexterous manipulation, grounds tasks in documented experimental SOPs, or measures per-step execution precision with continuous metrics.
A comparison is provided in Table~\ref{tab:comparison}.

\begin{table*}[t]
\centering
\caption{\textbf{Comparison with representative benchmarks.}
\cmark\,=\,supported, \pmark\,=\,partial, \xmark\,=\,not supported.
\textbf{Hum.}: humanoid embodiment;
\textbf{Dex.}: dexterous-hand manipulation;
\textbf{Lab}: scientific laboratory setting;
\textbf{SOP}: tasks derived from documented experimental procedures;
\textbf{Hier.}: multi-level task hierarchy;
\textbf{Constr.}: procedural or precondition constraints;
\textbf{Bi.}: bimanual manipulation;
\textbf{Instr.}: laboratory instrument interaction;
\textbf{Tool}: tool-mediated manipulation;
\textbf{Mat.}: experimental materials (liquids, powders, reagents);
\textbf{Step}: step-level evaluation;
\textbf{Prec.}: per-step precision metrics.
}
\label{tab:comparison}
\setlength{\tabcolsep}{2pt}
\footnotesize
\begin{tabular}{@{}l c cc cccc cccc cc@{}}
\toprule
\textbf{Benchmark}
& \textbf{Domain}
& \multicolumn{2}{c}{\textbf{Embodiment}}
& \multicolumn{4}{c}{\textbf{Protocol Grounding}}
& \multicolumn{4}{c}{\textbf{Scientific Manip.}}
& \multicolumn{2}{c}{\textbf{Evaluation}} \\
\cmidrule(lr){3-4}
\cmidrule(lr){5-8}
\cmidrule(lr){9-12}
\cmidrule(lr){13-14}
&
& \textbf{Hum.}
& \textbf{Dex.}
& \textbf{Lab}
& \textbf{SOP}
& \textbf{Hier.}
& \textbf{Constr.}
& \textbf{Bi.}
& \textbf{Instr.}
& \textbf{Tool}
& \textbf{Mat.}
& \textbf{Step}
& \textbf{Prec.} \\
\midrule
RLBench~\cite{rlbench}
& Tabletop
& \xmark & \xmark
& \xmark & \xmark & \xmark & \xmark
& \xmark & \xmark & \cmark & \xmark
& \xmark & \xmark \\
RoboCasa~\cite{robocasa}
& Household
& \xmark & \xmark
& \xmark & \xmark & \xmark & \xmark
& \xmark & \xmark & \xmark & \xmark
& \xmark & \xmark \\
ManiSkill~3~\cite{maniskill3}
& General
& \cmark & \cmark
& \xmark & \xmark & \xmark & \xmark
& \pmark & \xmark & \cmark & \xmark
& \xmark & \xmark \\
LIBERO~\cite{libero}
& Tabletop
& \xmark & \xmark
& \xmark & \xmark & \xmark & \xmark
& \xmark & \xmark & \xmark & \xmark
& \xmark & \xmark \\
RoboTwin~2~\cite{robotwin2}
& General
& \xmark & \xmark
& \xmark & \xmark & \xmark & \xmark
& \cmark & \xmark & \cmark & \xmark
& \xmark & \xmark \\
GenieSim~3~\cite{geniesim3}
& Humanoid
& \cmark & \pmark
& \xmark & \xmark & \xmark & \xmark
& \cmark & \xmark & \xmark & \xmark
& \xmark & \xmark \\
Factory~\cite{factory}
& Industrial
& \xmark & \xmark
& \xmark & \xmark & \xmark & \xmark
& \xmark & \xmark & \cmark & \xmark
& \xmark & \xmark \\
DexJoCo~\cite{dexjoco}
& General
& \xmark & \cmark
& \xmark & \xmark & \xmark & \xmark
& \cmark & \xmark & \cmark & \xmark
& \xmark & \xmark \\
\midrule
LabUtopia~\cite{labutopia}
& Chem.\ Lab
& \xmark & \xmark
& \cmark & \xmark & \cmark & \pmark
& \xmark & \cmark & \cmark & \cmark
& \pmark & \xmark \\
Chemistry3D~\cite{chemistry3d}
& Chem.\ Lab
& \xmark & \xmark
& \cmark & \xmark & \xmark & \xmark
& \xmark & \pmark & \xmark & \cmark
& \xmark & \xmark \\
AutoBio~\cite{autobio}
& Bio.\ Lab
& \xmark & \pmark
& \cmark & \xmark & \cmark & \pmark
& \cmark & \cmark & \cmark & \cmark
& \pmark & \pmark \\
MATTERIX~\cite{matterix}
& Chem.\ Lab
& \xmark & \xmark
& \cmark & \xmark & \xmark & \xmark
& \xmark & \cmark & \xmark & \cmark
& \xmark & \xmark \\
\midrule
\textbf{\ours (Ours)}
& Chem.\ Lab
& \cmark & \cmark
& \cmark & \cmark & \cmark & \cmark
& \cmark & \cmark & \cmark & \cmark
& \cmark & \cmark \\
\bottomrule
\end{tabular}

\end{table*}

\section{Simulation}
\label{sec:simulator}

Simulating laboratory tasks is more demanding than simulating home or tabletop tasks.
Home and tabletop benchmarks mainly evaluate whether the manipulated object reaches its goal pose.
A laboratory benchmark must additionally model \emph{functional instrument states}, \emph{quantitative tolerances} specified by experimental protocols, and \emph{material-level physics} that govern experimental observables.
Solid weighing, the most demanding task in our suite, exemplifies all three requirements.
The powder must be transferred as measurable mass, not merely rendered as a visual effect.
The analytical balance must update its reading as powder is deposited on the pan.
The task is scored against a protocol tolerance of $\pm 0.001$\,g.
Thus, a completed trajectory is valid only if it produces a scientifically meaningful mass measurement.

The \ours simulation is built from three components.
First, we construct laboratory scenes with \emph{functional assets and physics simulation}, including particle-based powder modeling and closed-loop instrument readouts (Sec.~\ref{sec:scenes_assets}).
Second, we develop an \emph{SOP-to-simulation pipeline} that converts documented experimental procedures into executable simulation tasks with built-in scoring (Sec.~\ref{sec:sop_pipeline}).
Third, we build a \emph{data collection infrastructure} around the Tianyi humanoid robot with dexterous-hand teleoperation (Sec.~\ref{sec:embodiment_data}).
Together, these provide the simulation foundation for \ours.
The task suite, evaluation hierarchy, and perturbation conditions are defined in Sec.~\ref{sec:benchmark}.

\subsection{Laboratory Scenes, Assets, and Physics Simulation}
\label{sec:scenes_assets}

\paragraph{Scene layout.}
We construct laboratory scenes centered on a fume hood serving as the primary workbench, with the required instruments and containers arranged within it.
All assets are sourced from ArtVIP~\cite{artvip} and follow its 3D modeling specification, with high-quality meshes, realistic textures, and physically accurate collision geometry.
This ensures that each asset is suitable for both visual rendering and physics simulation.
Two categories of scene layouts are provided.
\emph{Digital-twin scenes} faithfully reproduce real organic chemistry workstation arrangements, preserving the exact spatial configuration for sim-to-real evaluation.
\emph{Procedural scenes}, inspired by procedural generation in embodied AI~\cite{procthor}, use parameterized layouts in which instrument placement is randomized under axis-aligned bounding box (AABB) constraints.
These constraints guarantee no overlap in the top-down projection, enabling systematic generalization evaluation without spatial interference.
All scenes are designed to match real laboratory layouts.
Bench height, inter-instrument spacing, and camera viewpoints follow human operator conventions while remaining compatible with the Tianyi humanoid's reach envelope and head-mounted perception.

\paragraph{Functional assets.}
The asset library comprises over 30 objects in three categories: containers (round-bottom flasks, beakers, graduated cylinders, reagent bottles), tools (spatulas, funnels, micropipettes, magnetic stir bars), and instruments (analytical balances with weighing pans, heating devices).
Each asset is modeled as a collision-ready mesh with physically accurate inertial properties.
Instrument assets are further equipped with articulated joint definitions and \emph{measurable state variables} that mirror the quantities monitored during real experiments.
Articulated components such as balance windshield doors and tare buttons expose joint positions that serve as preconditions and postconditions in step definitions.
For example, windshield doors are modeled with prismatic-joint constraints so that a pinch-and-slide grasp on the door handle drives the joint along its guide track.

\paragraph{Powder physics and weighing simulation.}
Solid weighing requires simulating powder as a dry granular material that can be scooped, transferred, and deposited with physically meaningful mass.
The built-in particle system in Isaac Sim~\cite{isaacsim} is optimized for fluid-like materials.
In our experiments, it exhibited excessive fluidity that failed to reproduce the discrete, heap-forming behavior of dry chemical powders.
It also incurred prohibitive computational cost when large numbers of particles interacted with containers and the balance pan.
We therefore adopt a \emph{micro-element approach}: each powder grain is represented as a small rigid body, and contacts among particles and surrounding objects are resolved by the PhysX~\cite{isaacsim} collision solver.
We further tune the PhysX solver parameters and GPU memory allocation to maintain real-time performance and stable powder accumulation during weighing.
When particles land on the analytical balance pan, the balance computes the total deposited mass in real time and displays it on a simulated digital readout (Figure~\ref{fig:assets_b}).
This directly mirrors the workflow of a real analytical balance.
The closed-loop path from particle-level physics through mass computation to instrument display provides the physical grounding for the continuous precision metric $e_m$ (Sec.~\ref{sec:metrics}).
Critically, a weighing task succeeds not when powder visually appears on the pan, but when the mass reading falls within the protocol-specified tolerance ($\pm 0.001$\,g).
This ensures that evaluation is tied to the same quantitative standard used in real experiments.

\subsection{SOP-to-Simulation Pipeline}
\label{sec:sop_pipeline}

\begin{figure*}[t]
  \centering
  \begin{subfigure}[t]{0.48\textwidth}
    \centering
    \includegraphics[width=\linewidth]{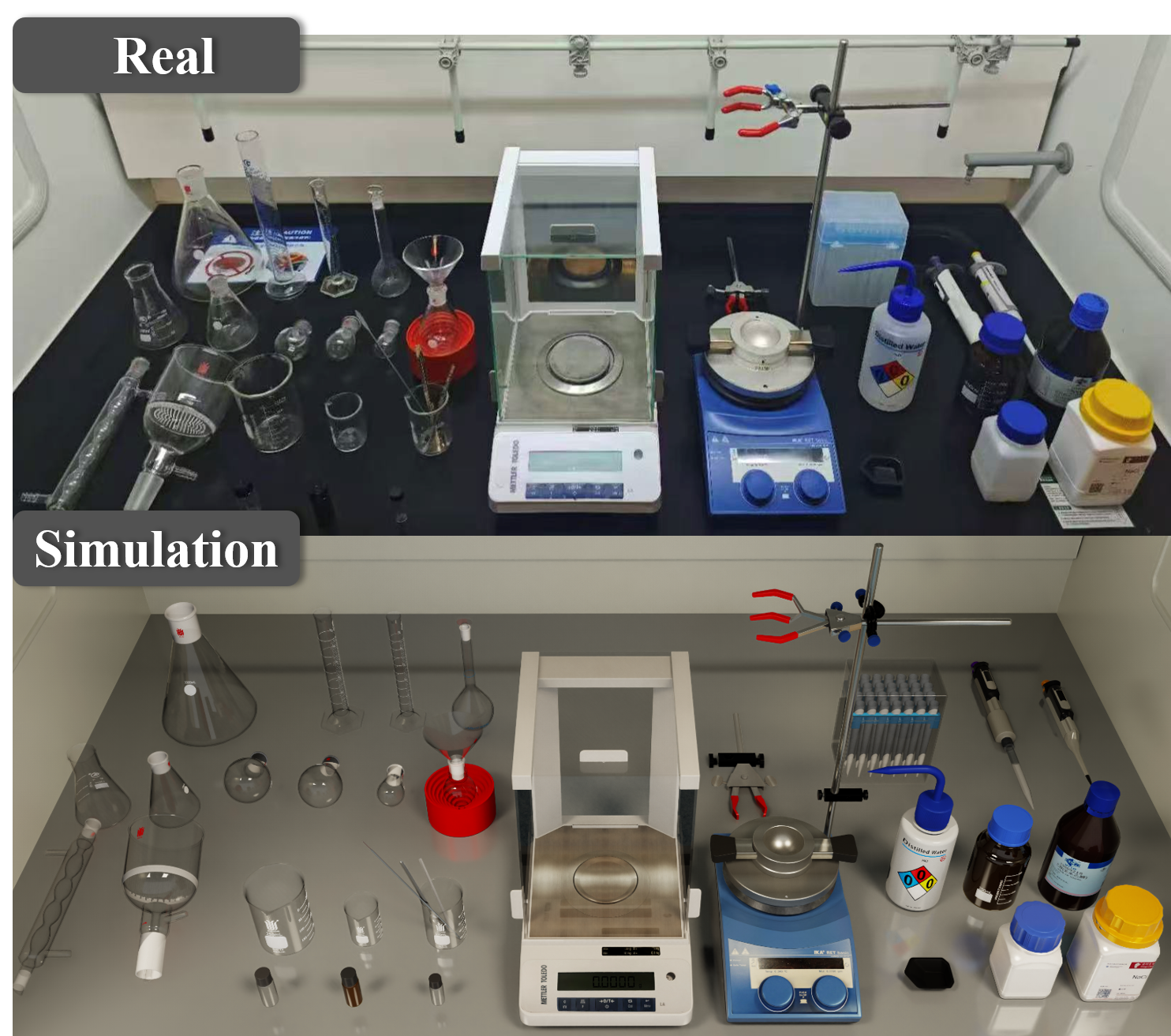}
    \caption{Representative functional assets.}
    \label{fig:assets_a}
  \end{subfigure}
  \hfill
  \begin{subfigure}[t]{0.48\textwidth}
    \centering
    \includegraphics[width=\linewidth]{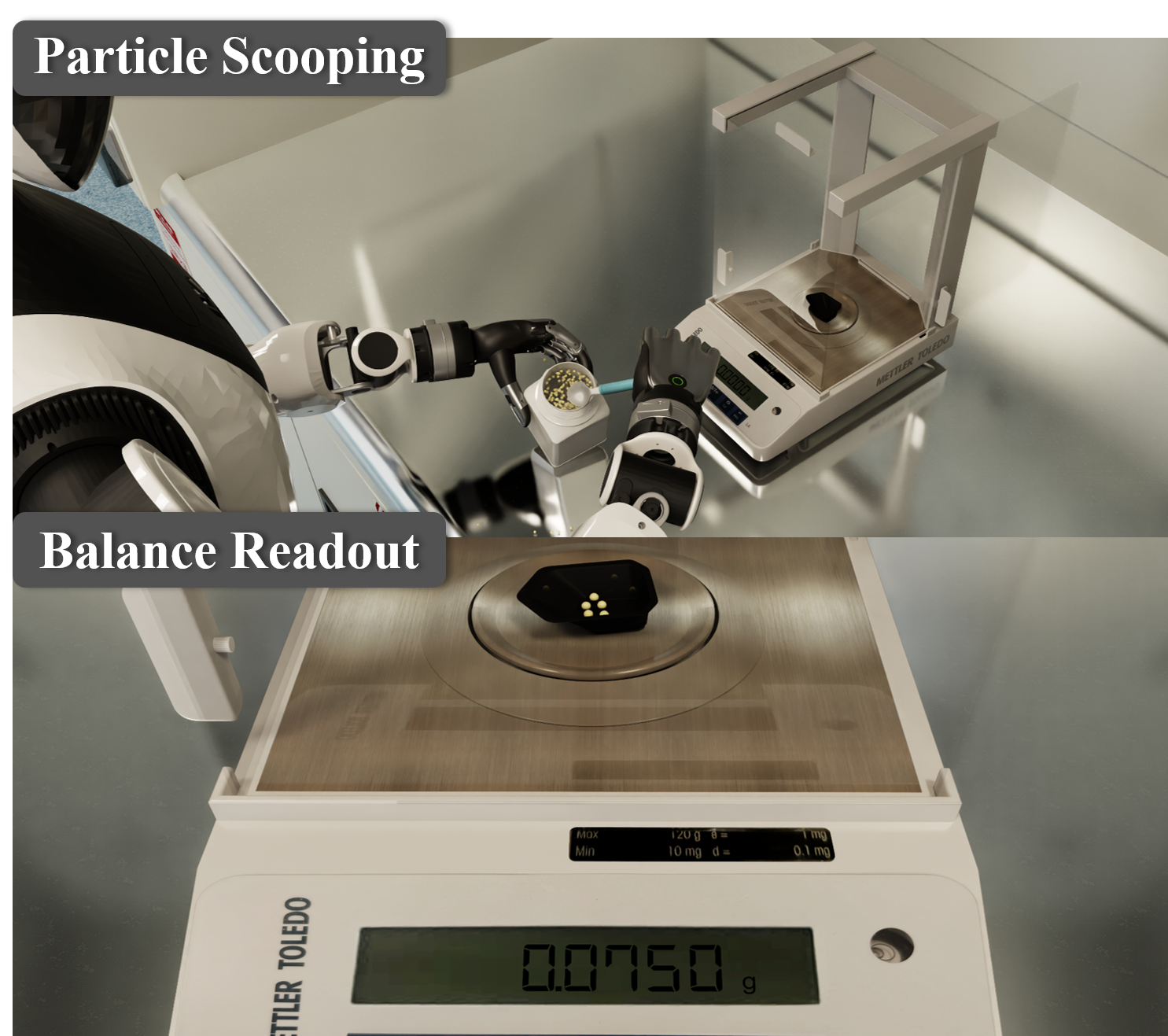}
    \caption{Particle-based weighing with digital readout.}
    \label{fig:assets_b}
  \end{subfigure}

  \vspace{4pt}

  \begin{subfigure}[t]{0.72\textwidth}
    \centering
    \includegraphics[width=\linewidth]{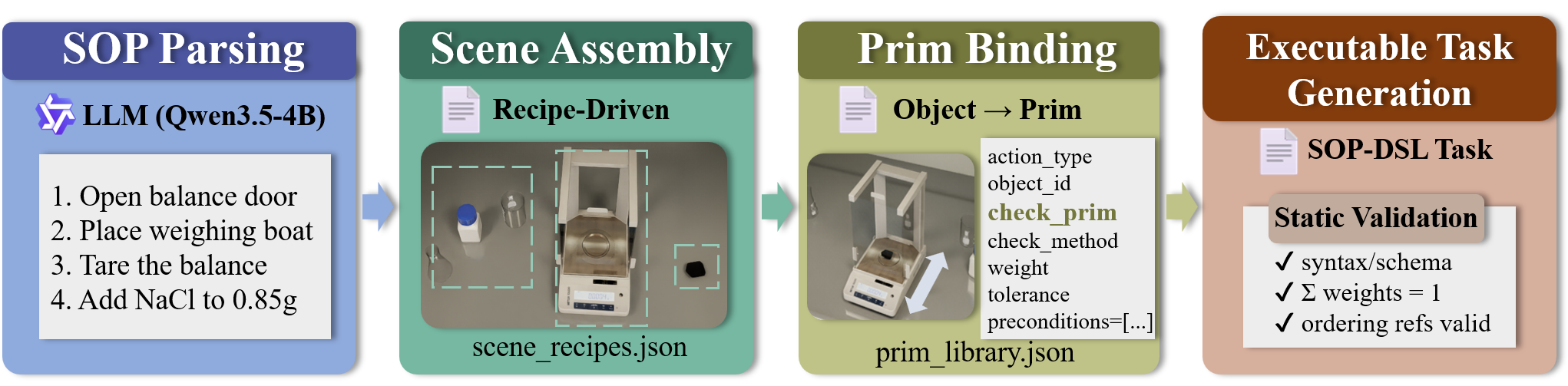}
    \caption{SOP-to-simulation pipeline.}
    \label{fig:pipeline}
  \end{subfigure}
  \hfill
  \begin{subfigure}[t]{0.26\textwidth}
    \centering
    \includegraphics[width=\linewidth]{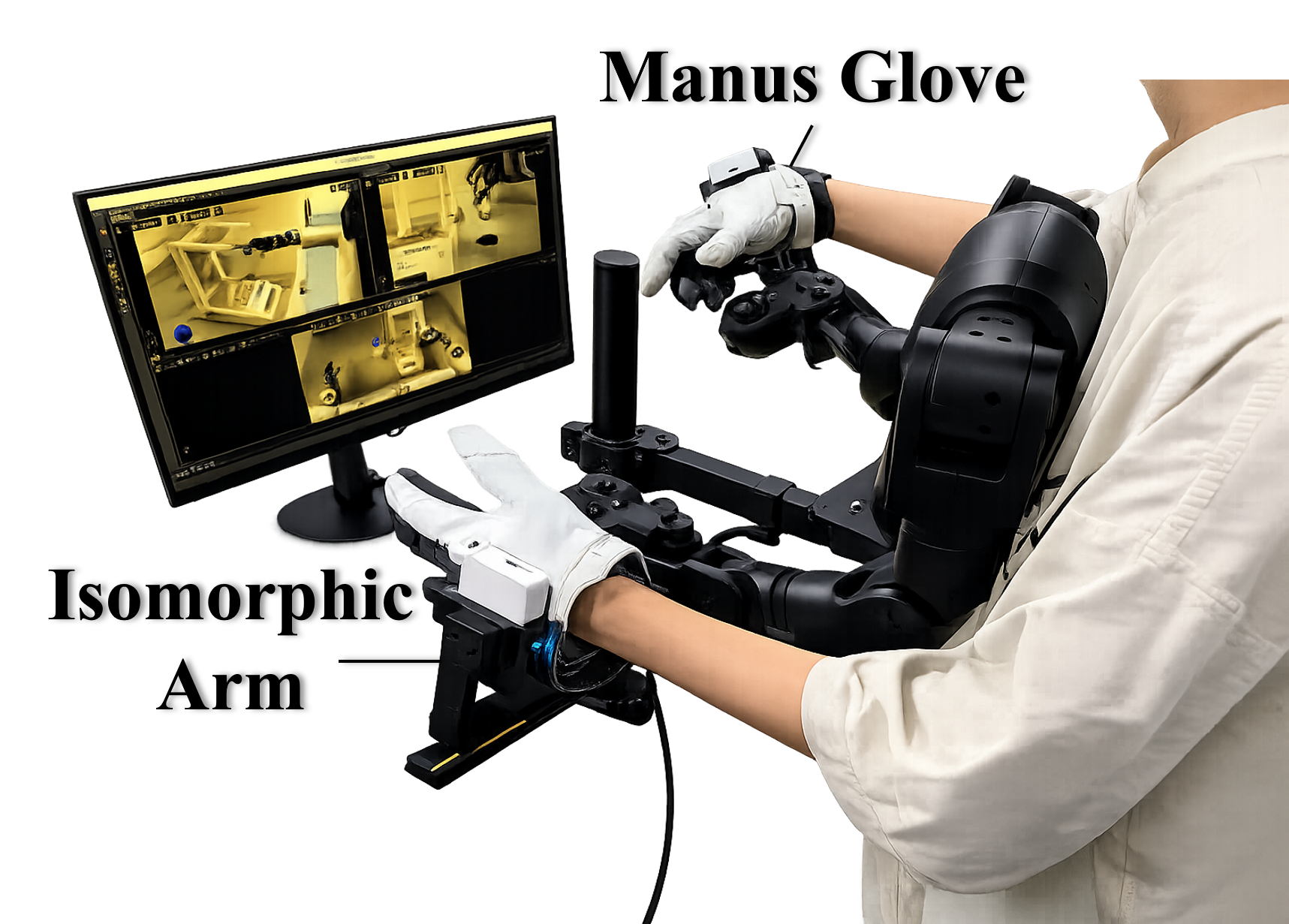}
    \caption{Teleoperation system.}
    \label{fig:teleop}
  \end{subfigure}

  \caption{\textbf{\ours simulation foundation.}
  \textbf{(a)}~Over 30 functional assets spanning containers, tools, and instruments.
  \textbf{(b)}~Rigid-body particles deposited on the balance pan; the digital readout displays the accumulated mass in real time.
  \textbf{(c)}~The SOP-to-simulation pipeline converts documented procedures into executable simulation tasks scored against the protocol specification.
  \textbf{(d)}~The operator wears Manus gloves and uses isomorphic arms to control the Tianyi humanoid.}
  \label{fig:sim_overview}
\end{figure*}

A central design goal of \ours is that every simulation task should be \emph{derived from} a real experimental procedure, not hand-crafted by engineers.
This ensures that the tasks, tolerances, and success criteria in the benchmark faithfully reflect real laboratory practice.
We realize this through a four-stage, LLM-assisted pipeline (Figure~\ref{fig:pipeline}) that converts a documented standard operating procedure (SOP) into an executable simulation task scored against the protocol specification.
We use this pipeline to construct the solid-weighing task at the core of the current benchmark and design it to be reapplied as further SOPs are added.

\paragraph{Stage 1: SOP parsing.}
The pipeline takes an experimental procedure as structured text input.
A large language model (Qwen3.5-4B), guided by a domain-specific prompt template, decomposes the procedure into manipulation-relevant steps.
Purely cognitive or observational instructions (\eg, ``record the mass reading,'' ``wait until the reading stabilizes'') are discarded because they have no checkable physical effect.
Each retained step $s_i$ is represented as a structured tuple:
\begin{equation}
  s_i = (a_i,\; o_i,\; m_i,\; w_i,\; T_i,\; P_i),
  \label{eq:sop_step}
\end{equation}
where $a_i$ is the manipulation action (\eg, \texttt{open\_door}, \texttt{place}, \texttt{weigh}) and $o_i$ is the target object.
$m_i$ is the check method specifying how completion is detected (\eg, joint state, spatial containment, mass reading).
$w_i$ is the step's weight in the task score.
$T_i$ is the quantitative tolerance bound to $m_i$ (\eg, $\pm 0.001$\,g for mass).
$P_i$ lists prerequisite steps that enforce execution order.
The prompt constrains the model to use action, object, and method terms from a fixed vocabulary.
This ensures that every generated step maps to a detector implemented in the simulator.
Step weights are normalized to sum to one across the procedure.
Because the language model can produce malformed outputs, all results pass through the validator in Stage~4 and are manually corrected where necessary.

\paragraph{Stage 2: Scene assembly.}
Rather than synthesizing layouts from scratch, the pipeline assembles each task scene from a curated recipe keyed by experiment type.
A recipe specifies the base laboratory scene and the functional assets the procedure requires.
Asset poses are pre-extracted from a real organic chemistry workstation.
The pipeline instantiates these assets into the base scene so that the same SOP always yields the same canonical layout.
Spatial, lighting, and texture variations are introduced separately as controlled perturbation conditions (Sec.~\ref{sec:benchmark}).

\paragraph{Stage 3: Prim binding and task specification.}
Each step's target object $o_i$ is a logical name (\eg, \texttt{balance\_door\_right}).
The pipeline resolves it to a concrete USD prim path through a static object-to-prim library.
This binding connects the abstract step to the specific simulator entity whose state the detector reads at run time.
For example, a balance door is bound to its prismatic joint for a joint-state check; the weighing pan is bound for a spatial-containment check.
The pipeline emits two configuration files.
The \emph{task configuration} records the scene path and robot initial pose.
The \emph{protocol specification} is a lightweight TOML-based representation (SOP-DSL) encoding the ordered, weighted step list with tolerances and bound prim paths.
These artifacts serve as both demonstration targets (Sec.~\ref{sec:embodiment_data}) and ground-truth specifications for evaluation (Sec.~\ref{sec:benchmark}).

\paragraph{Stage 4: Validation and executable task generation.}
Before a task enters data collection, the protocol specification is checked by a static validator.
The validator verifies three properties: (i)~syntactic and schema correctness, (ii)~that step weights sum to one, and (iii)~that all ordering references point to declared steps.
Failures are reported at step level, enabling targeted correction without regenerating the entire specification.
The validated pair of task configuration and protocol specification constitutes the executable task consumed by downstream data collection and evaluation.

\subsection{Tianyi Humanoid Teleoperation and Data Collection}
\label{sec:embodiment_data}

\paragraph{Humanoid embodiment.}
All demonstrations in \ours are collected using the Tianyi humanoid robot, a dual-arm upper-body platform with dexterous hands.
Each side provides a 7-DOF arm and a 6 active-DOF dexterous hand (BrainCo Revo 2).
The torso height and arm reach envelope are designed to match a standing adult operator's workspace.

\paragraph{Manus-glove and isomorphic-arm teleoperation.}
Demonstration trajectories are collected through a teleoperation system co-developed with the Tianyi team.
Recent dexterous teleoperation platforms~\cite{anyteleop} have demonstrated effective human-to-robot hand mapping.
Our system adopts a similar philosophy with two components tailored to laboratory manipulation.
The \emph{Manus glove} captures the operator's finger joint angles and hand pose in real time, providing direct kinematic mapping to the Tianyi dexterous hands.
The \emph{isomorphic-arm setup} mirrors the kinematic structure of the Tianyi arms, minimizing the human-to-robot mapping gap.
This enables the operator's arm motions to transfer naturally without mental retargeting.
This combination allows operators to perform contact-rich laboratory manipulations (weighing-boat placement, spatula scooping) with the same hand postures and arm trajectories used in a real laboratory.
Compared to indirect input devices, this substantially improves demonstration quality for precision-critical operations.
The teleoperation system refreshes at 90\,Hz.

\paragraph{Recorded modalities.}
Each demonstration records synchronized RGB observations from a head-mounted camera and two external cameras ($1280 \times 720$).
Proprioceptive states include arm joint positions ($2 \times 7$ dimensions) and hand joint angles ($2 \times 6$ dimensions), totaling 26 dimensions.

\section{Benchmark}
\label{sec:benchmark}

\begin{figure}[t!]
\centering
\includegraphics[width=\textwidth]{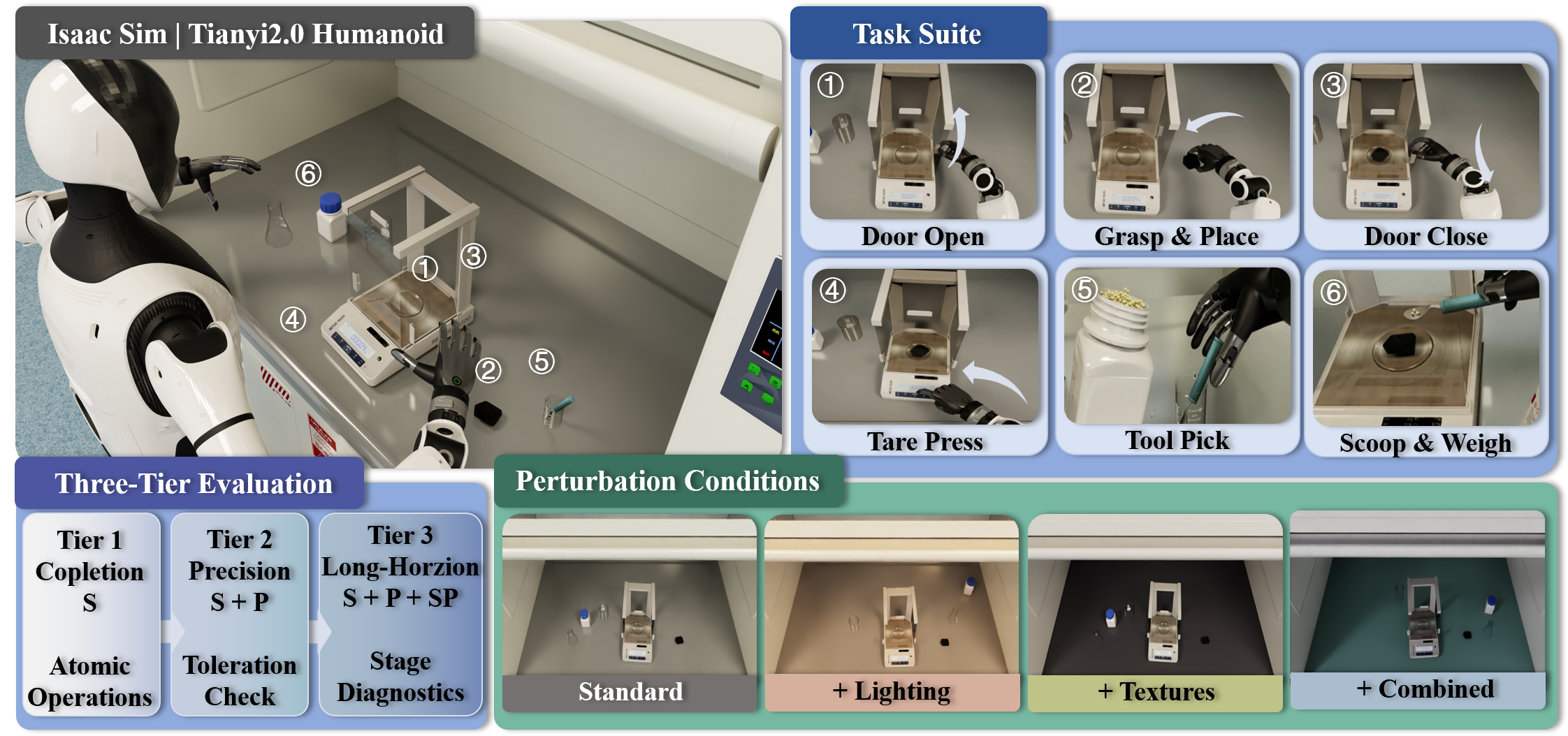}
\caption{\textbf{\ours benchmark overview.}
\emph{Top-left}: the simulation environment in Isaac Sim with the Tianyi humanoid.
\emph{Top-right}: the task suite spans six atomic operations (door open, door close, grasp \& place, tare press, tool pickup, and scoop \& weigh), covering discrete and sustained contact, single-arm and bimanual coordination, and instrument interaction.
\emph{Bottom-left}: the three-tier evaluation hierarchy progresses from binary task completion (Tier~1) through quantitative precision (Tier~2) to long-horizon stage diagnostics (Tier~3).
\emph{Bottom-right}: perturbation conditions (Standard, +Lighting, +Textures, and Combined) are applied uniformly across all tiers to assess robustness.}
\label{fig:benchmark}
\end{figure}

Task completion is a necessary but insufficient criterion for laboratory manipulation.
In chemistry, a robot may transfer powder onto a weighing pan (task complete) yet deliver a mass that exceeds the SOP-specified tolerance (experimentally invalid).
Existing manipulation benchmarks adopt binary task success and therefore cannot distinguish these two qualitatively different outcomes.

\ours addresses this gap with a structured evaluation framework organized along three axes.
We first describe the task suite derived from real organic chemistry procedures (Sec.~\ref{sec:tasks}).
We then introduce a three-tier evaluation hierarchy that progresses from binary completion through quantitative precision to long-horizon precision maintenance (Sec.~\ref{sec:hierarchy}).
Each tier adds an evaluation dimension that reveals failure modes invisible at the previous tier.
To assess robustness, all tasks are evaluated on procedural layouts (the Standard condition), with three perturbations layered on top: lighting, texture, and their combination (Sec.~\ref{sec:conditions}).
Finally, we define the metrics used at each tier (Sec.~\ref{sec:metrics}).

\subsection{Task Suite}
\label{sec:tasks}

\begin{figure*}[t]
  \centering
  \begin{subfigure}[t]{\textwidth}
    \centering
    \includegraphics[width=\linewidth]{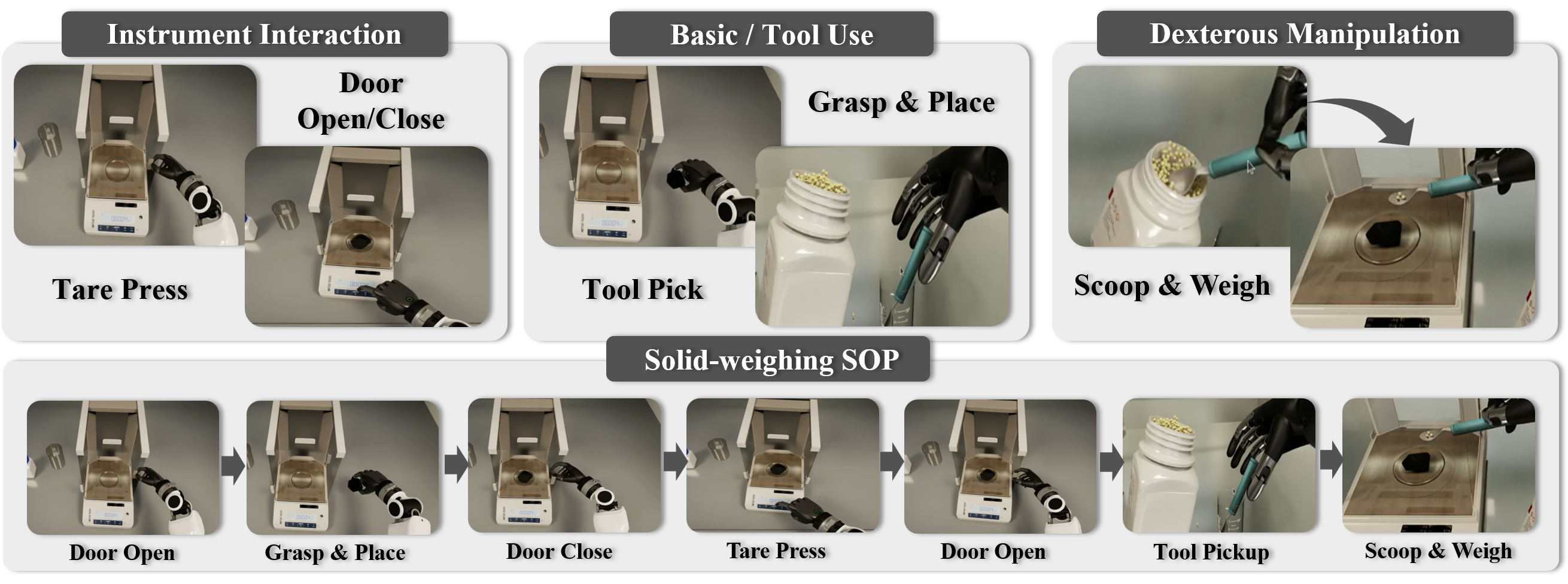}
    \caption{Atomic operations grouped by manipulation characteristics (\emph{top}) and the solid-weighing workflow (\emph{bottom}).}
    \label{fig:task_vis}
  \end{subfigure}

  \vspace{4pt}

  \begin{subfigure}[t]{\textwidth}
    \centering
    \includegraphics[width=\linewidth]{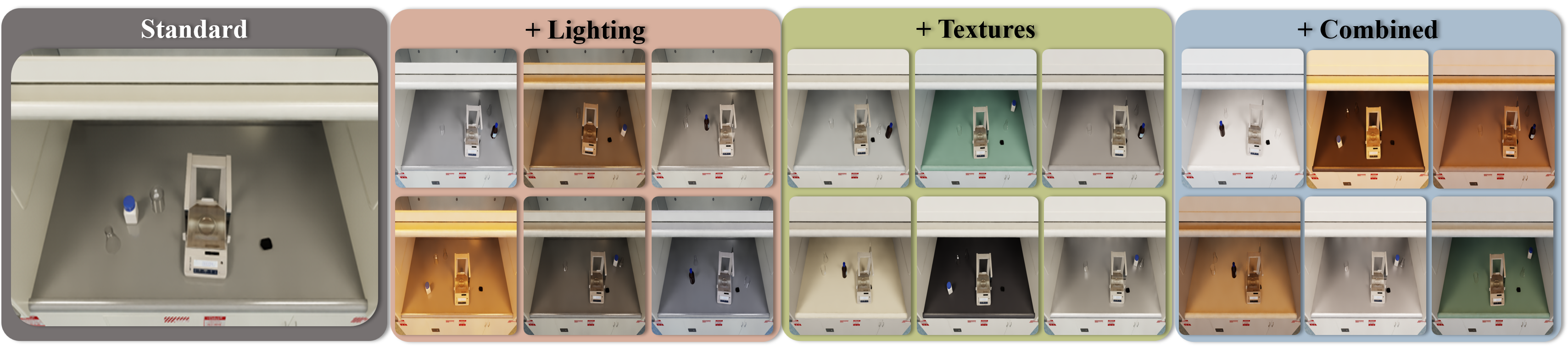}
    \caption{Evaluation conditions: Standard (procedural layouts), +Lighting, +Textures, and +Combined.}
    \label{fig:perturbation}
  \end{subfigure}

  \caption{\textbf{Solid-weighing task suite and evaluation conditions.}
  \textbf{(a)}~The task suite spans three manipulation categories (instrument interaction, basic tool use, and dexterous manipulation), with the solid-weighing procedure as a 7-step workflow (precision target 0.850\,$\pm$\,0.001\,g).
  \textbf{(b)}~All tiers are evaluated under four conditions with layered perturbations applied on top of procedural layouts.}
  \label{fig:benchmark_tasks}
\end{figure*}

\begin{table*}[t]
\centering
\caption{\textbf{\ours task suite.}
\textbf{Con.}: D\,=\,discrete, S\,=\,sustained, M\,=\,mixed.
\textbf{Arm}: 1\,=\,single, 2\,=\,bimanual.
\textbf{Dex.}: \cmark\,=\,required, \pmark\,=\,beneficial, \xmark\,=\,not required.
\textbf{Instr.}: instrument interaction.
\textbf{Prec.}: SOP precision target.
\textbf{Wt.}: per-step SOP weight in the Tier~3 workflow ($\sum\!=\!1$ across all 7 steps; door open contributes twice).
$\dagger$\,indicates precision evaluated at Tier~2; binary success only at Tier~1.}
\label{tab:task_suite}
\renewcommand{\arraystretch}{1.2}
\small
\setlength{\tabcolsep}{8pt}
\begin{tabular}{@{}l c c c c l c@{}}
\toprule
\textbf{Task} & \textbf{Contact} & \textbf{Arms} & \textbf{Dexterous} & \textbf{Instrument} & \textbf{Precision Target} & \textbf{Wt.} \\
\midrule
\multicolumn{7}{@{}l}{\textit{Single-arm atomic operations}} \\[3pt]
Grasp \& place      & D & 1 & \pmark & \xmark & $e_p \leq 15$\,mm$^\dagger$ & 0.15 \\
Door open            & D & 1 & \pmark & \cmark & ---                          & 0.10 \\
Door close           & D & 1 & \pmark & \cmark & ---                          & 0.10 \\
Tare press           & D & 1 & \cmark & \cmark & ---                          & 0.10 \\
Tool pickup          & S & 1 & \pmark & \xmark & ---                          & 0.10 \\
\midrule
\multicolumn{7}{@{}l}{\textit{Bimanual atomic operations}} \\[3pt]
Scoop \& weigh       & S & 2 & \cmark & \cmark & $e_m \leq 0.001$\,g$^\dagger$ & 0.35 \\
\midrule
\multicolumn{7}{@{}l}{\textit{Procedural workflow}} \\[3pt]
Solid weighing (7 steps) & M & 2 & \cmark & \cmark & $e_m \leq 0.001$\,g & 1.00 \\
\bottomrule
\end{tabular}
\end{table*}
All tasks in \ours are derived from the SOP-to-simulation pipeline described in Sec.~\ref{sec:sop_pipeline}.
Table~\ref{tab:task_suite} lists the complete task suite (six atomic operations and one multi-step procedural workflow), and Figure~\ref{fig:task_vis} visualizes them in the simulation workspace.
Each task is annotated along five manipulation characteristics that determine its placement in the evaluation hierarchy (Sec.~\ref{sec:hierarchy}):
\textbf{Contact type} (discrete or sustained),
\textbf{Arm configuration} (single-arm or bimanual),
\textbf{Dexterous-hand dependency} (required, beneficial, or not required),
\textbf{Instrument interaction} (whether the task involves a scientific instrument with measurable state changes), and
\textbf{Precision target} (whether the task has an SOP-specified quantitative tolerance, \eg, $\pm 0.001$\,g).

\subsection{Three-Tier Evaluation Hierarchy}
\label{sec:hierarchy}

Existing benchmarks commonly stratify tasks by the number of manipulation steps or environmental complexity~\cite{labutopia,robotwin2}, implicitly treating longer or harder-perturbed sequences as more challenging.
In scientific experiments, however, a single-step scooping operation with a $\pm 0.001$\,g tolerance can be more challenging than a five-step open-close sequence.
We therefore introduce a hierarchy distinguished not by step count but by the \emph{evaluation dimensions} applied to each tier.
Each successive tier adds a new dimension that reveals a failure mode invisible at the previous tier.

\paragraph{Tier 1: Task Completion.}
All six atomic operations are evaluated with \emph{binary success} only: did the intended state change occur?
Opening the balance door, pressing the tare button, grasping and placing a container, and scooping with a spatula are all Tier~1 tasks regardless of their physical complexity.
The evaluation asks only whether the operation was completed, not how precisely.
Tier~1 establishes a completion-rate baseline against which higher tiers reveal the precision gap.

\paragraph{Tier 2: Quantitative Precision.}
Two Tier~1 operations with SOP-specified quantitative tolerances are re-evaluated with additional \emph{continuous precision metrics}:
scoop \& weigh (mass error $e_m \leq 0.001$\,g, target 0.850\,g)
and grasp \& place (position error $e_p \leq 15$\,mm from the pan center).
Crucially, the same physical actions appear in both Tier~1 and Tier~2, but the evaluation criteria differ.
Tier~1 asks only ``did the transfer happen?'' (a binary 0/1 judgment on the end state).
Tier~2 further asks ``how accurately was it done?'' by measuring the \emph{continuous} deviation from the protocol target.
A policy may achieve high success at Tier~1 yet lower precision pass at Tier~2, because a fraction of completed transfers violate the quantitative tolerance.
This paired comparison directly quantifies how much binary success overestimates experimentally valid execution.
Tier~2 thus evaluates whether a policy possesses \emph{quantitative closed-loop control}, not merely contact ability.

\paragraph{Tier 3: Long-Horizon Precision.}
The complete solid-weighing workflow (7 steps, 6 distinct atomic operations, two stages: preparation and weighing) is evaluated with the full metric suite.
In addition to $S$ and $P$, Tier~3 introduces three new metrics.
The \emph{conditional precision pass rate} $P_{\text{cond}}$ (Eq.~\ref{eq:precision_cond}) measures precision exclusively among completed episodes.
\emph{Step progress} $\text{SP}$ (Eq.~\ref{eq:step_progress}) is the weighted fraction of completed protocol steps.
The \emph{stage-level breakdown} (Eq.~\ref{eq:sp_stage}) localizes failures to specific operational phases.
The defining challenge of Tier~3 is \emph{compounding precision error}.
Each preceding step introduces small pose and state perturbations that accumulate, degrading the precision achievable at later steps.
A policy with moderate isolated precision (Tier~2) may see that precision degrade substantially when the same operation is embedded in the full workflow (Tier~3).
Errors from door opening, boat placement, and taring compound and shift the workspace state before the precision-critical weighing step.
Stage-level diagnostics transform the evaluation from a single pass/fail verdict into a diagnostic profile that pinpoints which phase incurs the failure.

\subsection{Evaluation Conditions}
\label{sec:conditions}

Real laboratories are not static: lighting shifts with time of day, bench surfaces vary between workstations, and instrument placements change between experimental sessions.
A policy that achieves high precision by memorizing visual features or spatial positions from the training scene may fail when any of these factors change.
This is a well-known challenge that has motivated domain randomization in sim-to-real transfer~\cite{domainrand,openai_rubik}.
We therefore define a \emph{Standard} test condition that represents a spatial-distribution shift relative to the training layouts.
Three perturbation conditions progressively layer additional visual variation on top of this baseline.
Following COLOSSEUM~\cite{colosseum}, lighting and texture are evaluated as separate axes rather than merged into a single ``visual'' condition.
This enables attribution of precision loss to specific visual factors.

\paragraph{Standard condition (procedural layouts).}
The training scenes are replaced by procedural layouts (Sec.~\ref{sec:scenes_assets}) in which instrument and object placements are randomized within SOP-derived spatial constraints.
Approximately 50 procedural layouts are pre-generated; each test episode samples one at random.
Lighting and textures are held at default values.
All other perturbation conditions are layered on top of this Standard baseline.

\paragraph{+Lighting perturbation.}
Building on the Standard baseline, the lighting is additionally randomized.
Color temperature is uniformly sampled from 3000--7000\,K, intensity is scaled by a factor in $[0.4, 1.6]$, and each light source's direction is perturbed by $\pm 60^{\circ}$ along the elevation and azimuth axes.
Textures remain at default.

\paragraph{+Texture perturbation.}
Building on the Standard baseline, the bench surface is replaced by one of 10 laboratory-plausible materials and the background wall by one of 5 materials.
Materials are defined via OmniPBR shader parameters (diffuse color, roughness, metallic).
Lighting remains at default.

\paragraph{+Combined perturbation.}
All three perturbation axes (procedural layout, randomized lighting, and randomized textures) are applied simultaneously, simulating realistic deployment conditions where multiple environmental factors change concurrently.

Each tier is evaluated under all four conditions: Standard and the three layered perturbations above (Figure~\ref{fig:perturbation}).
This produces a $3 \times 4$ evaluation matrix (Tier $\times$ Condition).
The matrix reveals both the per-tier precision gap and the marginal contribution of each perturbation axis.

\subsection{Metrics}
\label{sec:metrics}

Metrics are introduced incrementally across tiers; each tier's metric set is a strict superset of the previous tier's.

\paragraph{Task success rate ($S$).}
Applied to all tiers.
For each episode $i$, $S_i = 1$ if the terminal goal state is achieved within the time limit, and $S_i = 0$ otherwise.
We report the mean success rate $S = \frac{1}{N}\sum_{i=1}^{N} S_i$ across $N$ episodes.

\paragraph{Continuous precision error and precision pass rate ($P$).}
Applied to Tier~2 and Tier~3.
For tasks with quantitative targets, we define:
\begin{align}
    e_m &= |m_{\text{actual}} - m_{\text{target}}| \quad \text{(g, mass error)}, \label{eq:em} \\
    e_p &= |p_{\text{actual}} - p_{\text{target}}| \quad \text{(mm, position error)}, \label{eq:ep}
\end{align}
where $e_m$ is the mass error for weighing tasks and $e_p$ is the positional error for placement tasks.
We denote by $e_i$ the precision error for episode $i$ ($e_i = e_m$ or $e_p$ depending on the task) and by $S_i \in \{0, 1\}$ its binary success indicator.
The \emph{precision pass rate} $P$ is the fraction of episodes in which the task is both completed and the precision error falls within the SOP-specified tolerance:
\begin{equation}
    P = \frac{1}{N} \sum_{i=1}^{N} \mathds{1}[S_i = 1 \;\wedge\; e_i \leq \tau],
    \label{eq:precision_pass}
\end{equation}
where $\tau = T_i$ is the tolerance bound to the check method in the SOP-DSL task definition (Eq.~\ref{eq:sop_step}).
The gap $S - P$ directly quantifies the degree to which binary success overestimates experimentally valid execution.

To disentangle completion failures from precision failures, we additionally define the \emph{conditional precision pass rate}:
\begin{equation}
    P_{\text{cond}} = \frac{\sum_{i=1}^{N} \mathds{1}[S_i = 1 \;\wedge\; e_i \leq \tau]}{\sum_{i=1}^{N} \mathds{1}[S_i = 1]},
    \label{eq:precision_cond}
\end{equation}
which measures precision quality exclusively among completed episodes.
$P_{\text{cond}}$ is essential for cross-tier comparisons (Sec.~\ref{sec:exp_main}).
Comparing isolated precision (Tier~2) against the same operation in a long-horizon workflow (Tier~3) via $P$ alone would conflate genuine precision degradation with earlier-step failures that prevent the policy from reaching the precision-critical step.

\paragraph{Step progress and stage-level breakdown.}
Applied to Tier~3 only.
\emph{Step progress} $\text{SP} \in [0, 1]$ is the SOP-weighted fraction of protocol steps whose check methods pass before the episode terminates:
\begin{equation}
    \text{SP} = \sum_{j=1}^{K} w_j \cdot \mathds{1}[\text{step}_j \text{ passes}],
    \label{eq:step_progress}
\end{equation}
where $w_j$ is the step weight from the SOP-DSL (Eq.~\ref{eq:sop_step}) and $K$ is the number of protocol steps.
Weights encode each step's scientific importance.
The weighing step (0.35) dominates over place-weigh-boat (0.15) and door/tare operations (0.10 each), so SP reflects progress toward experimentally valid completion rather than raw step count.
For multi-stage workflows, per-stage step progress restricts the sum in Eq.~\ref{eq:step_progress} to steps within stage $k$ and renormalizes by the stage's total weight:
\begin{equation}
    \text{SP}^{(k)} = \frac{\sum_{j \in \mathcal{S}_k} w_j \cdot \mathds{1}[\text{step}_j \text{ passes}]}{\sum_{j \in \mathcal{S}_k} w_j},
    \label{eq:sp_stage}
\end{equation}
where $\mathcal{S}_k \subset \{1, \dots, K\}$ is the set of step indices belonging to stage $k$.
This localizes failures to specific operational phases (\eg, preparation completed but weighing collapsed).

\section{Experiments}
\label{sec:exp}


\subsection{Setup}
\label{sec:exp_setup}

\paragraph{Tasks.}
We evaluate four atomic operations that currently have trained policies (Table~\ref{tab:task_suite}).
At Tier~1, door open, door close, and tare press are assessed with binary success rate~$S$ across all three baselines.
These tasks involve instrument interaction with increasing demands on fine contact control.
At Tier~2, grasp \& place is evaluated with both $S$ and precision pass rate~$P$ (position error $e_p \leq 15$\,mm from pan center) for ACT.
Evaluation of the remaining tasks (tool pickup, scoop \& weigh) and the complete solid-weighing workflow (Tier~3) is ongoing.
Each task-condition pair is evaluated over 50 episodes $\times$ 3 random seeds (150 trials per entry).

\paragraph{Baselines.}
We select three baselines spanning imitation learning~\cite{robomimic,cliport} and generalist robot policies~\cite{rt1,rt2,octo,openvla}.
All baselines operate on the Tianyi humanoid with dexterous hands (Sec.~\ref{sec:embodiment_data}).

\begin{itemize}
    \setlength\itemsep{0.1em}

    \item \textbf{ACT}~\cite{act_ref}:
    Action Chunking with Transformers, a transformer-based imitation-learning method designed for bimanual manipulation.
    Receives RGB observations and proprioceptive states; predicts chunked action sequences.

    \item \textbf{Diffusion Policy (DP)}~\cite{dp_ref}:
    a diffusion-based imitation-learning method that models the action distribution as a conditional denoising process.
    Receives the same observation modalities as ACT.

    \item \textbf{$\pi_0$}~\cite{pi0_ref}:
    a vision-language-action (VLA) foundation model that receives task instructions and visual observations, outputting low-level actions via flow matching.
\end{itemize}

\paragraph{Training protocol.}
All baselines are trained on 100 teleoperation demonstrations per task.
Demonstrations are collected via the Manus-glove and isomorphic-arm teleoperation system (Sec.~\ref{sec:embodiment_data}) in digital-twin laboratory scenes.
Lighting, surface textures, and physical parameters remain fixed at default values during training.
At test time, the Standard condition introduces a spatial-distribution shift relative to the training layouts (Sec.~\ref{sec:conditions}).
The three visual perturbations (lighting, texture, and their combination) are layered on top.

Observations comprise RGB images from a head-mounted camera ($1280 \times 720$), resized to $640 \times 480$ for ACT and DP, and $224 \times 224$ for $\pi_0$.
For grasp \& place, two external cameras are added to mitigate arm and hand self-occlusion.
Proprioceptive states include arm joint positions and hand joint angles (26 dimensions total; Sec.~\ref{sec:embodiment_data}).
ACT and DP are trained from scratch; $\pi_0$ is fine-tuned from a pretrained checkpoint.
All baselines share identical demonstration data and action spaces (26-dimensional joint positions).
$\pi_0$ additionally conditions on a task instruction string, consistent with its VLA design.

\subsection{Results and Analysis}
\label{sec:exp_main}

Table~\ref{tab:results} presents the experimental results. We first analyze Tier~1 task completion across all baselines, then examine the Tier~2 precision gap for grasp \& place, and finally evaluate robustness under visual perturbations.

\paragraph{Task-level discrimination.}
Within Tier~1, the three tasks exhibit a clear difficulty gradient.
Door open achieves the highest success rates across all baselines (47.3--56.7\%), followed by door close (6.7--40.7\%).
Tare press is near zero for all methods (ACT 2.0\%, DP and $\pi_0$ both 0.0\%).
This gradient reflects increasing demands on fine contact control.
Tare pressing demands precise fingertip contact on a small button.
The near-zero success rates across all three methods suggest that this type of single-finger manipulation remains an open challenge for current imitation learning and VLA methods, even with 100 training demonstrations.

\paragraph{Baseline differentiation.}
No single baseline dominates across all tasks.
ACT achieves the highest success on door open (56.7\%).
$\pi_0$ leads on door close (40.7\%), outperforming both ACT (24.7\%) and DP (6.7\%) by a substantial margin.
The near-zero tare press results across all baselines point to a shared limitation of current methods rather than an architecture-specific weakness.

\paragraph{Precision gap: completion $\neq$ precision.}
Table~\ref{tab:results}(b) presents the Tier~2 precision analysis for ACT on grasp \& place.
The same episodes are evaluated under both binary success $S$ (``did the boat reach the pan?'') and precision pass rate $P$ (``was it placed within 15\,mm of the pan center?'').
ACT achieves $S = 5.3 \pm 0.9$\,\% but only $P = 3.3 \pm 1.9$\,\%, yielding a gap of $2.0$ percentage points.
Although the absolute numbers are low due to the difficulty of the task itself, the gap confirms that a non-trivial fraction of completed episodes violate the placement tolerance.
Among the 8 completed episodes across all seeds, only 5 satisfy the placement tolerance ($P_{\text{cond}} \approx 62.5\%$), indicating that task completion alone does not guarantee precision.
This supports the core finding that binary success overestimates experimentally valid execution.

\subsection{Robustness Analysis}
\label{sec:exp_robust}

To probe the robustness of the VLA baseline, we evaluate $\pi_0$ on Door open under the four conditions defined in Sec.~\ref{sec:conditions}.
The trained model from Sec.~\ref{sec:exp_main} is kept unchanged.

\begin{table*}[t]
\centering
\caption{\textbf{Experimental results under the Standard condition.}
\textbf{(a)}~Tier~1 task success rate $S$\,(\%), mean$\pm$std across 3 seeds.
\textbf{(b)}~Tier~2 precision analysis for ACT on grasp \& place, per-seed. $P$: precision pass rate ($e_p \leq 15$\,mm).
\textbf{(c)}~Robustness of $\pi_0$ on door open under four conditions, per-seed.}
\label{tab:results}
\footnotesize
\setlength{\tabcolsep}{5pt}
\renewcommand{\arraystretch}{1.15}

\vspace{4pt}

\begin{subtable}[t]{\textwidth}
\centering
\begin{tabular*}{\textwidth}{@{\extracolsep{\fill}}lccc@{}}
\toprule
\textbf{Task} & \textbf{ACT} & \textbf{DP} & $\boldsymbol{\pi_0}$ \\
\midrule
Door open            & $56.7 \pm 3.4$ & $49.3 \pm 2.5$ & $47.3 \pm 7.4$ \\
Door close           & $24.7 \pm 0.9$ & $6.7 \pm 2.5$  & $40.7 \pm 3.8$ \\
Tare press           & $2.0 \pm 0.0$  & $0.0 \pm 0.0$  & $0.0 \pm 0.0$  \\
\bottomrule
\end{tabular*}
\caption{Tier~1: task success $S$\,(\%)}
\end{subtable}

\vspace{6pt}

\begin{subtable}[t]{0.38\textwidth}
\centering
\begin{tabular}{@{}lcc@{}}
\toprule
 & \textbf{$\boldsymbol{S}$\,(\%)} & \textbf{$\boldsymbol{P}$\,(\%)} \\
\midrule
Seed 1 & 4.0 & 2.0 \\
Seed 2 & 6.0 & 6.0 \\
Seed 3 & 6.0 & 2.0 \\
\midrule
$\mu \pm \sigma$ & $5.3{\pm}0.9$ & $3.3{\pm}1.9$ \\
\bottomrule
\end{tabular}
\caption{Tier~2: ACT on grasp \& place}
\end{subtable}
\hfill
\begin{subtable}[t]{0.58\textwidth}
\centering
\begin{tabular}{@{}lcccc@{}}
\toprule
 & \textbf{Std.} & \textbf{+Light} & \textbf{+Tex.} & \textbf{+Comb.} \\
\midrule
Seed 1 & 48 & 48 & 42 & 32 \\
Seed 2 & 56 & 36 & 52 & 44 \\
Seed 3 & 38 & 40 & 46 & 44 \\
\midrule
$\mu \pm \sigma$ & $47.3{\pm}7.4$ & $41.3{\pm}5.0$ & $46.7{\pm}4.1$ & $40.0{\pm}5.7$ \\
\bottomrule
\end{tabular}
\caption{Robustness: $\pi_0$ on door open}
\end{subtable}

\end{table*}

\paragraph{Key findings.}
$\pi_0$ is relatively insensitive to isolated visual perturbations.
Performance under lighting-only ($41.3{\pm}5.0$\,\%) and texture-only ($46.7{\pm}4.1$\,\%) conditions remains close to the Standard baseline ($47.3{\pm}7.4$\,\%).
The combined condition produces the largest drop ($40.0{\pm}5.7$\,\%, $-7.3$\,pp), exceeding the effect of any single perturbation alone.
This suggests that compounded visual perturbations have a greater impact on VLA performance than isolated ones.

\section{Conclusion}
\label{sec:conclusion}

We presented \ours, to our knowledge the first benchmark for humanoid dexterous manipulation in precision-critical laboratory environments.
Built from real organic chemistry workstations through real-to-sim reconstruction, \ours includes more than 30 functionally faithful assets that collectively cover the core operations of routine organic chemistry experiments.
The benchmark integrates articulated laboratory instruments, particle-based powder physics, and closed-loop instrument readouts, enabling realistic evaluation from physical manipulation to quantitative measurement.
It further introduces a precision-aware evaluation protocol that jointly measures task completion, experimental precision, and long-horizon execution.
We benchmarked three representative policies across multiple laboratory manipulation tasks.
The results reveal a consistent precision gap: policies that successfully complete manipulation tasks often fail to satisfy the quantitative tolerances required by laboratory protocols.
This finding highlights a fundamental disconnect between task completion and experimentally valid manipulation, which cannot be captured by conventional binary success metrics.
Beyond providing a benchmark, \ours offers a foundation for developing humanoid robots capable of scientific experimentation.
Many laboratory operations, including fine-force powder transfer, milligram-level weighing, and contact-rich instrument interaction, fundamentally require dexterity, coordination, and human-like embodiment.
We hope \ours will serve as both an evaluation platform and a catalyst for future research on embodied AI for laboratory automation.
While this work focuses on organic chemistry, we plan to extend the benchmark to broader scientific disciplines and more complex experimental workflows.

\paragraph{Limitations and future work.}
The current release represents \textbf{version 0} of \ours, and we explicitly scope it as an initial foundation rather than a finished benchmark.
(i)~The experimental evaluation covers four representative atomic operations and one Tier~2 precision task.
Evaluation of the remaining tasks (tool pickup, scoop \& weigh) and the complete solid-weighing workflow (Tier~3) is ongoing and will be included in subsequent versions.
(ii)~The task suite currently focuses on solid weighing in organic chemistry.
Future versions will extend to liquid handling, apparatus assembly, and multi-step synthetic procedures, broadening the range of chemistry operations and dexterous capabilities under evaluation.
(iii)~All experiments are conducted in simulation; systematic sim-to-real transfer and real-world validation are planned as a critical next milestone.
(iv)~The benchmark currently supports a single humanoid platform (Tianyi); multi-embodiment support, including alternative humanoid and dexterous-hand configurations, will enable cross-platform comparison.
These limitations define a clear development roadmap.
We will iteratively expand the scope of the task, complete the evaluation hierarchy, and validate sim-to-real transfer, progressing \ours toward a comprehensive benchmark for humanoid laboratory manipulation.
\noindent
\paragraph{Acknowledgment.} This work was supported by the New Generation Artificial Intelligence-National Science and Technology Major Project (Grant No.2025ZD0122603 and No.2025ZD0122604). This work was also supported by Beijing Innovation Center of Humanoid Robotics (X-Humanoid).

{
    \small
    \bibliographystyle{unsrtnat}
    \bibliography{main}
}


\end{document}